\documentclass[runningheads]{llncs}
\usepackage{comment}
\usepackage{graphicx}
\usepackage{tabularx}
\usepackage{hyperref}
\usepackage{color}
\usepackage{booktabs}
\usepackage{amsfonts, amsmath, amssymb}
\usepackage[thicklines]{cancel}
\usepackage{multirow, multicol, graphicx}
\usepackage[table,xcdraw]{xcolor}
\usepackage{threeparttable}
\usepackage{marvosym}
\begin{document}
\newcommand{\wuzirui}[2]{\textcolor{red}{(#1)}\textcolor{blue}{#2}}
\newcommand{\yl}[1]{\textcolor{cyan}{\textbf{YL}: #1}}
\newcommand{\zknote}[1]{\textcolor{red}{\textbf{ZK}:(#1)}}
\newcommand{\ssec}[1]{\vspace{1mm}\noindent\textbf{#1.}}

\title{MARS: An Instance-aware, Modular and Realistic Simulator for Autonomous Driving}

\def\CICAISubNumber{12}  

\titlerunning{MARS}
%
\author{Zirui Wu\inst{1,2} \and
Tianyu Liu\inst{1,3} \and
Liyi Luo\inst{1,4}\and
Zhide Zhong\inst{1,5}\and
Jianteng Chen\inst{1,5}\and
Hongmin Xiao\inst{1,6}\and
Chao Hou\inst{1,7}\and
Haozhe Lou\inst{1,8}\and
Yuantao Chen\inst{1,9}\and\\
Runyi Yang\inst{1,10}\and
Yuxin Huang\inst{1,5}\and
Xiaoyu Ye\inst{1,5}\and\\
Zike Yan\inst{1}\and
Yongliang Shi\inst{1}\and
Yiyi Liao\inst{11}\and
Hao Zhao\inst{1}\thanks{Corresponding author. \email{zhaohao@air.tsinghua.edu.cn}.\\
Sponsored by Tsinghua-Toyota Joint Research Fund (20223930097) and Baidu Inc. through Apollo-AIR Joint Research Center.} 
}
\authorrunning{Z. Wu et al.}
\institute{$^1$ AIR, Tsinghua University; $^2$ System Thurst, HKUST(GZ); $^3$ HKUST; \\ $^4$ McGill University; $^5$ Beijing Institute of Technology; $^6$ National University of Singapore; $^7$ HKU; $^8$ University of Wisconsin Madison; $^9$ Xi’an University of Architecture and Technology; $^{10}$ Imperial College London; $^{11}$ Zhejiang University \vspace{-5mm}
 }
%


\maketitle              

\begin{abstract}
Nowadays, autonomous cars can drive smoothly in ordinary cases, and it is widely recognized that realistic sensor simulation will play a critical role in solving remaining corner cases by simulating them. To this end, we propose an autonomous driving simulator based upon neural radiance fields (NeRFs). Compared with existing works, ours has three notable features: (1) \textbf{Instance-aware}. Our simulator models the foreground instances and background environments separately with independent networks so that the static (e.g., size and appearance) and dynamic (e.g., trajectory) properties of instances can be controlled separately. (2) \textbf{Modular}. Our simulator allows flexible switching between different modern NeRF-related backbones, sampling strategies, input modalities, etc. We expect this modular design to boost academic progress and industrial deployment of NeRF-based autonomous driving simulation. (3) \textbf{Realistic}. Our simulator set new state-of-the-art photo-realism results given the best module selection. Our simulator will be \textbf{open-sourced} while most of our counterparts are not. Project page:  \url{https://open-air-sun.github.io/mars/}.

\end{abstract}


\section{Introduction}
Autonomous driving~\cite{geiger_vision_2013,hu_planning-oriented_2023,zheng_steps_2023,li_lode_2023,10160470,jin_adapt_2023} is arguably the most important application of modern 3D scene understanding~\cite{chen_pq-transformer_2022,tian_vibus_2022} techniques. Nowadays, Robotaxis can run in big cities with up-to-date HD maps, handling everyday driving scenarios smoothly. However, once a corner case that lies out of the distribution of an autonomous driving algorithm happens on the road unexpectedly, the lives of passengers are put at risk. The dilemma is that while we need more training data about corner cases, collecting them in the real world usually means danger and high expenses. To this end, the community believes that photorealistic simulation~\cite{li_aads_2019,chen_geosim_2021,yang_unisim_2023,fu_panoptic_2022} is a technical path of great potential. If an algorithm can experience enormous corner cases in a simulator with a small sim-to-real gap, the performance bottleneck of current autonomous driving algorithms can be potentially addressed. 

Existing autonomous driving simulation methods have their own limitations. CARLA~\cite{dosovitskiy_carla_2017} is a widely used sensor simulator based upon traditional graphics engines, whose realism is restricted by asset modeling and rendering qualities. AADS~\cite{li_aads_2019} also exploits traditional graphics engines but demonstrates impressive photorealism using well-curated assets. On the other hand, GeoSim~\cite{chen_geosim_2021} introduces a data-driven scheme for realistic simulation by learning an image enhancement network. Flexible asset generation and rendering can be achieved through image composition with promisingly good geometry and realistic appearance.

In this paper, we take advantage of the realistic rendering ability of NeRFs for autonomous driving simulation. Training data captured from real-world environments guarantees a small sim-to-real gap. Several works also exploit NeRFs to model cars~\cite{niemeyer_giraffe_2021} and static backgrounds~\cite{fu_panoptic_2022} in outdoor environments. However, the inability to model complex dynamic scenes that are composed of both moving objects and static environments limits their practical use for real-world sensor simulation. Recently, Neural Scene Graph (NSG)~\cite{ost_neural_2021} decomposes dynamic scenes into learned scene graphs and learns latent representations for category-level objects. However, its multi-plane-based representation for background modeling cannot synthesize images under large viewpoint changes. 

To be specific, our central contribution is the very first \textbf{open-source} NeRF-based modular framework for photorealistic autonomous driving simulation. The proposed pipeline models foreground instances and background environments in a decomposed fashion. Different NeRF backbone architectures and sampling methods are incorporated in a unified manner with multi-modal inputs supported. The best module combination of the proposed framework achieves state-of-the-art rendering performance on public benchmarks with large margins, indicating photorealistic simulation results.

\begin{figure}[t]
    \centering
    \includegraphics[width=0.9\textwidth]{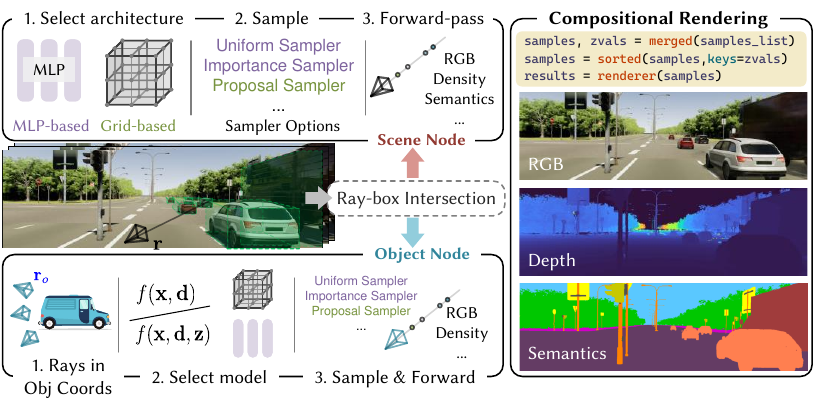}
    \caption{\textbf{Pipeline}. \textbf{Left}: We first calculate the ray-box intersection of the queried ray $\textbf{r}$ and all visible instance bounding boxes $\{\mathcal{B}_{ij}\}$. For the background node, we directly use the selected scene representation model and the chosen sampler to infer point-wise properties, as in conventional NeRFs. For the foreground nodes, the ray is first transformed into the instance frame as $\textbf{r}_o$ before being processed through foreground node representations (Sec.~\ref{sec:scene-representation}). \textbf{Right}: All the samples are composed and rendered into RGB images, depth maps, and semantics (Sec.~\ref{sec:rendering}).}
    \label{fig:main-figure}
    \vspace{-6mm}
\end{figure}

\vspace{-2mm}
\section{Method}
\vspace{-2mm}
\ssec{Overview}
As illustrated in~Fig.~\ref{fig:main-figure}, we aim to provide a modular framework for constructing compositional neural radiance fields, where realistic sensor simulation can be conducted for outdoor driving scenes.
A large unbounded outdoor environment with plenty of dynamic objects is taken into consideration.

The input to the system consists of a set of RGB-images $\{\mathcal{I}_i\}^N$ (captured by vehicle-side or roadside sensors), sensor poses $\{\mathcal{T}_i\}^N$ (calculated using IMU/GPS signals), and object tracklets (including 3D bounding boxes $\{\mathcal{B}_{ij}\}^{N\times M}$, categories $\{\mathtt{type}_{ij}\}^{N\times M}$, and instance IDs $\{\mathtt{idx}_{ij}\}^{N\times M}$). $N$ is the number of input frames and $M$ is the number of tracked instances $\{\mathcal{O}_j\}^M$ across the whole sequence. An optional set of depth maps $\{\mathcal{D}_i\}^N$ and semantic segmentation masks $\{\mathcal{S}_i\}^N$ can also be adopted as extra supervision signals during training.
By constructing a compositional neural field, the proposed framework can simulate realistic sensor perception signals (including RGB images, depth maps, semantic segmentation masks, etc.) at given sensor poses. Instance editing on object trajectories and appearances is also supported.

\ssec{Pipeline} Our framework model each foreground instance and the background node compositionally.  As shown in Fig.~\ref{fig:main-figure}, when querying properties (RGB, depth, semantics, etc.) of a given ray $\mathbf{r}$, we first calculate its intersection with all visible objects' 3D bounding boxes to get the entering and leaving distances $[t_\mathtt{in}, t_\mathtt{out}]$. Afterward, both the background node (Fig.~\ref{fig:main-figure} left-top) and the foreground object nodes (Fig.~\ref{fig:main-figure} left-bottom) are queried, where each node samples a set of 3D points and uses its specific neural representation network to obtain point properties (RGB, density, semantics, etc.). Specifically, to query foreground nodes, we convert the ray origins and directions from world space into instance frames according to the object tracklets. Finally, all the ray samples from background and foreground nodes are composed and volume-rendered to produce pixel-wise rendering results (Fig.~\ref{fig:main-figure} right, Sec.~\ref{sec:rendering}).

We observe that the nature of background nodes (typically unbounded large-scale scenes) differs from the object-centric foreground nodes, while current works~\cite{kundu_panoptic_2022,ost_neural_2021} in sensor simulation use unified NeRF models. Our framework provides a flexible and open-sourced framework that supports different design choices of scene representations for background and foreground nodes and can easily incorporate new state-of-the-art methods of static scene reconstruction and object-centric reconstructions.

\subsection{Scene Representation}
\label{sec:scene-representation}

We decompose the scene into a large-scale unbounded NeRF (as the background node) and multiple object-centric NeRFs (as independent foreground nodes). Conventionally, a neural radiance field maps a given 3D point coordinate $\mathbf{x} = (x,y,z), \mathbf{x} \in \mathbb{R}^3$ and a 2D viewing direction $\mathbf{d}\in\mathbb{S}^2$ to its radiance $\mathbf{c}$ and volume density $\sigma$ shown in Eq.~\ref{eq:nerf-field}. Based upon this seminal representation, many variants have been proposed for different purposes, so we take a modular design.
\begin{equation}
\label{eq:nerf-field}
    f (\mathbf{x}, \mathbf{d}) = (\mathbf{c}, \sigma): [\mathbb{R}^3, \mathbb{S}^2] \to [\mathbb{R}^3, \mathbb{R}^+]
\end{equation}
The challenge of modeling unbounded background scene photo-realistically lies in accurately representing far regions, so we utilize the unbounded scene warping~\cite{barron_mip-nerf_2022} to contract the far region. For foreground nodes, we support both the code-conditioned representation $f(\mathbf{x}, \mathbf{d}, \mathbf{z})= (\textbf{c}, \sigma)$ ($\mathbf{z} \in \mathbb{R}^k$ denotes the instance-wise latent code) and the conventional ones, which will be explained as follows.

\subsubsection{Architectures.} In our modular framework, we support various NeRF backbones, which can be roughly categorized into two hyper-classes: MLP-based methods~\cite{mildenhall_nerf_2020,barron_mip-nerf_2021,barron_mip-nerf_2022}, or grid-based methods that store spatially-variant features in their hash grid voxel vertices~\cite{muller_instant_2022,tancik_nerfstudio_2023}. Although these architectures differ from each other in details, they follow the same high-level formulation of Eq.~\ref{eq:nerf-field} and are capsuled in modules under a unified interface in MARS.

While the MLP-based representations are simple in mathematical form, we give a formal exposition of grid-based methods.
The specific implementation of a multi-resolution feature grid $\{\mathcal{G}_\theta^{l}\}^L_{l=1}$ has layer-wise resolutions $\displaystyle R_l:=\lfloor R_\text{min}\cdot b^l\rfloor, b = \exp\left(\frac{\ln R_{\max{}} - \ln R_{\min{}}}{L-1}\right)$, where $R_{\min{}}, R_{\max{}}$ are the coarsest and the finest resolution~\cite{yu_monosdf_2022,muller_instant_2022}. The coordinates $\mathbf{x}$ are first scaled to each resolution before being processed by the ceiling and flooring operations to $\lceil \mathbf{x}\cdot R_l\rceil, \lfloor\mathbf{x}\cdot R_l\rfloor$ and hashed to obtain table indexes~\cite{muller_instant_2022}. The extracted feature vectors are then tri-linearly interpolated and decoded through a shallow MLP.
\begin{equation}
    (\mathbf{c}, \sigma) = f_\theta\left(\mathtt{interp}(\mathtt{hash\_and\_lookup}(\mathbf{x}, \{\mathcal{G}_\theta^{l}\}^L_{l=1})), \mathbf{d}\right).
\end{equation}

\ssec{Sampling}  We support various sampling strategies, including the recently proposed proposal network~\cite{barron_mip-nerf_2022}, which distills a density field from a radiance-free NeRF model to generate ray samples and other sampling schemes like coarse-to-fine sampling \cite{mildenhall_nerf_2020} or uniform sampling \cite{fridovich-keil_k-planes_2023} for flexibility.


\ssec{Foreground Nodes}
For rendering foreground instances, we first transform the projected rays into per-instance coordinate space and then infer the object-centric NeRFs in each instance-wise canonical space.
The default setting of our framework uses code-conditioned models that exploit latent codes to encode instance features and \textbf{shared} category-level decoders to encode class-wise priors, allowing the modeling of many long tracklets with compact memory usage. Meanwhile, the conventional ones without code conditions are also supported in our framework.
We detailed our modified foreground representation (denoted as `Ours' in Sec.~\ref{sec:experiments}) in supplementary materials.



\begin{figure}[htb]
    \centering
    \includegraphics[width=0.9\textwidth]{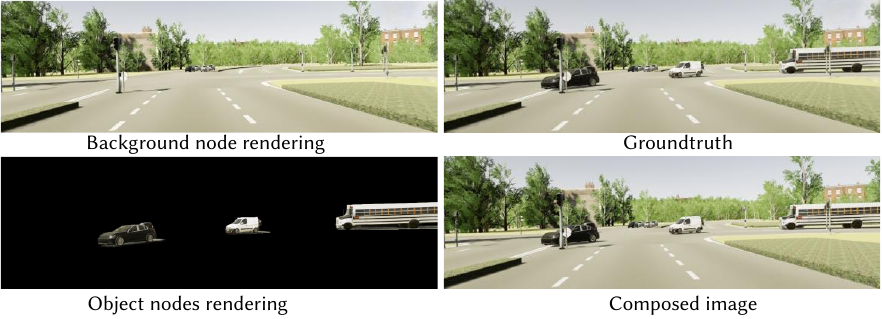}
    \caption{Illustration on the compositional rendering. Some of the static vehicles in the far region are considered as background objects. \vspace{-4mm}}
    \label{fig:compose}
\end{figure}

\subsection{Compositional Rendering}
\label{sec:rendering}
\begin{figure}[t]
    \centering
    \includegraphics[width=0.75\textwidth]{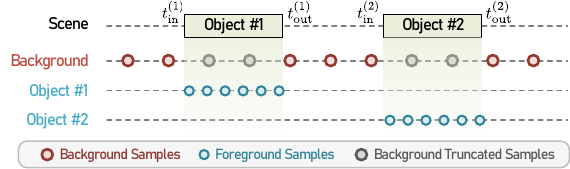}
    \caption{Visual demonstration on our \textbf{conflict-free} sampling process. We use uniform sampling in all nodes for illustration.\vspace{-4mm}}
    \label{fig:sampling}
\end{figure}

Figure~\ref{fig:compose} demonstrates the compositional rendering results. To render an image at a given camera pose $\mathcal{T}_i$, we cast a ray $\mathbf{r} = \mathbf{o} + t\mathbf{d}$ at each rendered pixel. For each ray $\mathbf{r}$, we first calculate the intersection interval $[t_\text{in}, t_\text{out}]$ with all visible foreground nodes $\mathcal{O}_{ij}$ (Fig.~\ref{fig:sampling}) and transform the samples $\{P_k^\text{obj-j}\}$ along the ray from world space into each foreground canonical space.
We also sample a set of 3D points along the ray ($\{P_k^\text{bg}\}$ as background samples. Samples in all nodes are first passed through their corresponding networks to obtain point-wise colors $\{\textbf{c}_k^{\text{bg, obj}}\}$, densities $\{\sigma_k^{\text{bg, obj}}\}$, and foreground semantic logits $\{\textbf{s}_k^{\text{bg}}\}$. 
Considering that the semantic properties of foreground samples are actually their category label, we create a one-hot vector as:
\begin{equation}
    \textbf{s}_k^{\text{obj-j}}[l] = \left\{ \begin{matrix}
        &\sigma_k^{\text{obj-j}} &\qquad \text{if } l = \text{category of j's instance}\\
        &0 &\text{otherwise}
    \end{matrix} \right., \text{for}\ l\ \text{in category}.
\end{equation}
To aggregate the point-wise properties, we sort all the samples by their ray distance in world space and use the standard volume rendering process to render pixel-wise properties:
\begin{align}
    &\hat{\textbf{c}}(\textbf{r}) = \sum_{P_i}T_i\alpha_i \textbf{c}_i + (1 - \texttt{accum})\cdot \textbf{c}_\text{sky},\ \ T_i = \exp(-\sum_{k=1}^{i-1}\sigma_k\delta_k), \label{eq:render}\\
   & \hat{d}(\textbf{r})\! =\! \sum_{P_i}T_i\alpha_i t_i + (1 - \texttt{accum})\cdot \texttt{inf}, \ \hat{\textbf{s}}(\textbf{r})\! =\! \sum_{P_i}T_i\alpha_i \textbf{s}_i +(1 - \texttt{accum})\cdot \textbf{s}_\text{sky},
\end{align}
where $P_i\in\mathtt{sorted}(\{P_i^\text{bg, obj}\}), \alpha_i=1-\exp(-\sigma_i\delta_i), \delta_i=t_{i+1}-t_i$, $\texttt{accum} = \sum_{P_i} T_i \alpha_i$ , $\textbf{c}_\text{sky}$ is the rendered color from the Sky model (Sec.\ref{sec:sky-and-removebg}), $\texttt{inf}$ is the upper bound distance, and $\textbf{s}_\text{sky}$ is the one-hot semantic logits of the $\texttt{sky}$ category.

\subsection{Towards Realistic Rendering}
\label{sec:sky-and-removebg}
\ssec{Sky Modeling}
In our framework, we support the usage of a sky model to deal with appearances at infinite distance, where an MLP-based spherical environment map~\cite{rematas_urban_2022} is leveraged to model the infinitely far regions that never intersect opaque surfaces:
\begin{equation}
    f_\text{sky}(\mathbf{d}) = \textbf{c}_\text{sky}: \mathbb{S}^2 \to \mathbb{R}^3
\end{equation}

However, na\"ively blending the sky color $\textbf{c}_\text{sky}$ with background and foreground rendering (Eq.~\ref{eq:render}) leads to potential inconsistency. Therefore, we introduce a BCE semantic regularization to alleviate this issue:
\begin{equation}
    \mathcal{L}_\text{sky} = \texttt{BCE}(1 - \texttt{accum}, \mathcal{S}_\text{sky}).
\label{eq:sky_loss}
\end{equation}

\subsubsection{Resolving Conflict Samples.}
\label{sec:resolving conflict}
\begin{figure}[ht]
\vspace{-4mm}
    \centering
    \includegraphics[width=0.8\textwidth]{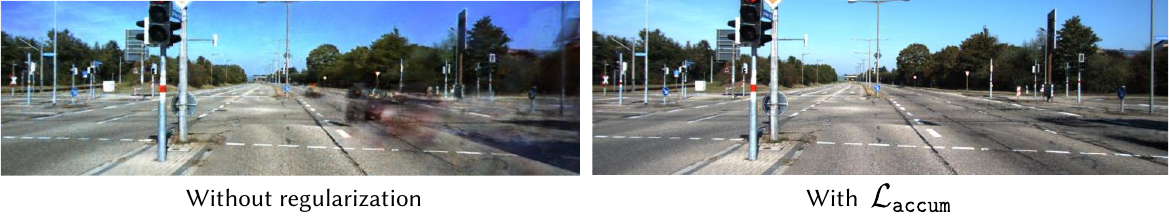}
    \vspace{-4mm}
    \caption{We show that the background truncated samples cause background-foreground ambiguity without our regularization.}
    \label{fig:no_reg}
\vspace{-5mm}
\end{figure}

Due to the fact that our background and foreground sampling are done independently, there is a chance that background samples fall within the foreground bounding box (Fig.~\ref{fig:sampling} Background Truncated Samples). The compositional rendering may mistakenly classify foreground samples as background (referred to later as background-foreground ambiguity). As a result, after removing the foreground instance, artifacts will emerge in the background area (Fig.~\ref{fig:no_reg}). Ideally, with sufficient multi-view supervision signal, the system can automatically learn to distinguish between foreground and background during the training process. However, for a data-driven simulator, obtaining abundant and high-quality multi-view images is challenging for users as vehicles move fast on the road. The ambiguity is \textbf{NOT} observed in NSG~\cite{ost_neural_2021} as NSG only samples a few points on the ray-plane intersections, and is unlikely to have much background truncated samples.

To address this issue, we devise a regularization term that minimizes the density sum of background truncated samples to minimize their influence during the rendering process as:
\begin{equation}
    \mathcal{L}_\texttt{accum} = \sum_{P_i^\text{(tr)}} \sigma_i,
\label{eq:acc_loss}
\end{equation}
where $\{P_i^\text{(tr)}\}$ denotes background truncated samples.


\subsection{Optimization}
\label{sec:optimization}

To optimize our system, we minimize the following objective function:
\begin{equation}
    \mathcal{L} = \lambda_1\mathcal{L}_\text{color} + \lambda_2\mathcal{L}_\text{depth} + \lambda_3\mathcal{L}_\text{sem} + \lambda_4\mathcal{L}_\text{sky} + \lambda_5\mathcal{L}_\texttt{accum},
\end{equation}
where $\lambda_{1-5}$ are weighting parameters. $\mathcal{L}_\text{sky}$ and $\mathcal{L}_\texttt{accum}$ are explained in Eq.~\ref{eq:sky_loss} and~\ref{eq:acc_loss}.

\noindent\textbf{Color Loss}: we adopt a standard MSE loss that minimizes the photo-metric errors as:
\begin{equation}
    \mathcal{L}_\text{color} = ||\textbf{c}(\textbf{r}) - \hat{\textbf{c}}(\textbf{r})||_2^2.
\end{equation}

\noindent\textbf{Depth Loss}:
We introduce a depth loss to address textureless regions and regions that are observed from sparse viewpoints. We have devised two strategies for supervising the geometry. Given depth data, we utilize a ray distribution loss derived from~\cite{deng_depth-supervised_2022}. On the other hand, if the depth data is not available, we utilize a mono-depth network and apply mono-depth loss following~\cite{yu_monosdf_2022}.
\begin{equation}
\mathcal{L}_{\text{depth}}=\left\{ \begin{matrix}
	&\mathcal{L}_{\text{sensor\_depth}} \qquad &\text{if depth data is available}\\
	&\mathcal{L}_{\text{mono\_depth}} &\text{if depth data is not available}\\
\end{matrix} \right. 
\end{equation}

\noindent\textbf{Semantic Losses}: we follow SemanticNeRF~\cite{zhi_-place_2021} and use a cross-entropy semantic loss $\mathcal{L}_\text{sem} = \texttt{CrossEntropy}(\textbf{s}(\textbf{r}), \mathcal{S}(\textbf{r}))$.


\section{Experiments}
\label{sec:experiments}
In this section, we provide extensive experimental results to demonstrate the proposed instance-aware, modular, and realistic simulator for autonomous driving.
We evaluate our method on scenes from the KITTI~\cite{geiger_vision_2013} dataset and the Virtual KITTI-2 (V-KITTI)~\cite{cabon_virtual_2020} dataset. In the following, we use \textbf{``our default setting''} to denote a grid-based NeRF with proposal sampler for the background node, and our modified category-level representation with coarse-to-fine sampler for foreground nodes.

\begin{table}[htbp]
  \vspace{-3mm}
  \centering
  \caption{Qunatitative results on image reconstruction task \& Comparisons on the settings with baseline methods. The dataset used for evaluation is KITTI.}
  \resizebox{0.8\textwidth}{!}{
    \begin{tabular}{cccccccc}
    \hline
          &  NeRF \cite{mildenhall_nerf_2020}  & NeRF+Time & NSG \cite{ost_neural_2021}  & PNF  \cite{kundu_panoptic_2022} & SUDS \cite{turki_suds_2023}  & Ours \\
    \hline
    PSNR $\uparrow$ &  23.34 & 24.18 & 26.66 & 27.48 & 28.31 & \textbf{29.06} \\
    SSIM $\uparrow$ &  0.662 & 0.677 & 0.806 & 0.870  & 0.876 & \textbf{0.885}  \\
    \hline
    Instance-aware & $\times$ & $\times$ & \checkmark & \checkmark & $\times$  & \checkmark\\
    Modular & $\times$ & $\times$ & $\times$ & $\times$ & $\times$  & \checkmark\\
    Open-sourced & \checkmark & - & \checkmark & $\times$ & \checkmark  & \checkmark\\
    \hline
    \end{tabular}%
    }
  \label{tab:recon-quant}%
  \vspace{-10mm}
\end{table}%


\begin{table}[ht]
\centering
\caption{Qunatitative results on novel view synthesis}
\resizebox{0.75\textwidth}{!}{%
\begin{tabular}{lccccccccc}
\hline
\multicolumn{1}{c}{} &
  \multicolumn{3}{c}{\textbf{KITTI-75\%}} &
  \multicolumn{3}{c}{\textbf{KITTI-50\%}} &
  \multicolumn{3}{c}{\textbf{KITTI-25\%}} \\ \cline{2-10} 
\multicolumn{1}{c}{} &
  PSNR $\uparrow$&
  SSIM $\uparrow$&
  LPIPS $\downarrow$&
  PSNR $\uparrow$&
  SSIM $\uparrow$&
  LPIPS $\downarrow$&
  PSNR $\uparrow$&
  SSIM $\uparrow$&
  LPIPS $\downarrow$\\ \hline
NeRF~\cite{mildenhall_nerf_2020} &
  18.56 &
  0.557 &
  0.554 &
  19.12 &
  0.587 &
  0.497 &
  18.61 &
  0.570 &
  0.510 \\
NeRF+Time &
  21.01 &
  0.612 &
  0.492 &
  21.34 &
  0.635 &
  0.448 &
  19.55 &
  0.586 &
  0.505 \\
NSG~\cite{ost_neural_2021} &
  21.53 &
  0.673 &
  0.254 &
  21.26 &
  0.659 &
  0.266 &
  20.00 &
  0.632 &
  0.281 \\
SUDS~\cite{turki_suds_2023} &
  22.77 &
  0.797 &
  0.171 &
  23.12 &
  \textbf{0.821} &
  \textbf{0.135} &
  20.76 &
  0.747 &
  0.198 \\
Ours &
  \textbf{24.23} &
  \textbf{0.845} &
  \textbf{0.160} &
  \textbf{24.00} &
  0.801 &
  0.164 &
  \textbf{23.23} &
  \textbf{0.756} &
  \textbf{0.177} \\
\rowcolor[HTML]{EFEFEF} 
\multicolumn{1}{c}{\cellcolor[HTML]{EFEFEF}} &
  {\color[HTML]{FF0000} +1.46} &
  {\color[HTML]{FF0000} +0.048} &
  {\color[HTML]{FF0000} -0.011} &
  {\color[HTML]{FF0000} +0.88} &
  {\color[HTML]{32CB00} -0.020} &
  {\color[HTML]{32CB00} +0.029} &
  {\color[HTML]{FF0000} +2.47} &
  {\color[HTML]{FF0000} +0.009} &
  {\color[HTML]{FF0000} -0.021} \\ \hline
\multicolumn{1}{c}{} &
  \multicolumn{3}{c}{\textbf{VKITTI-75\%}} &
  \multicolumn{3}{c}{\textbf{VKITTI-50\%}} &
  \multicolumn{3}{c}{\textbf{VKITTI-25\%}} \\ \cline{2-10} 
\multicolumn{1}{c}{} &
  PSNR $\uparrow$&
  SSIM $\uparrow$&
  LPIPS $\downarrow$&
  PSNR $\uparrow$&
  SSIM $\uparrow$&
  LPIPS $\downarrow$&
  PSNR $\uparrow$&
  SSIM $\uparrow$&
  LPIPS $\downarrow$\\ \hline
NeRF~\cite{mildenhall_nerf_2020} &
  18.67 &
  0.548 &
  0.634 &
  18.58 &
  0.544 &
  0.635 &
  18.17 &
  0.537 &
  0.644 \\
NeRF+Time &
  19.03 &
  0.574 &
  0.587 &
  18.90 &
  0.565 &
  0.610 &
  18.04 &
  0.545 &
  0.626 \\
NSG~\cite{ost_neural_2021} &
  23.41 &
  0.689 &
  0.317 &
  23.23 &
  0.679 &
  0.325 &
  21.29 &
  0.666 &
  0.317 \\
SUDS~\cite{turki_suds_2023} &
  23.87 &
  0.846 &
  0.150 &
  23.78 &
  0.851 &
  0.142 &
  22.18 &
  0.829 &
  0.160 \\
Ours &
  \textbf{29.79} &
  \textbf{0.917} &
  \textbf{0.088} &
  \textbf{29.63} &
  \textbf{0.916} &
  \textbf{0.087} &
  \textbf{27.01} &
  \textbf{0.887} &
  \textbf{0.104} \\
\rowcolor[HTML]{EFEFEF} 
\multicolumn{1}{c}{\cellcolor[HTML]{EFEFEF}} &
  {\color[HTML]{FF0000} +5.92} &
  {\color[HTML]{FF0000} +0.071} &
  {\color[HTML]{FF0000} -0.062} &
  {\color[HTML]{FF0000} +5.85} &
  {\color[HTML]{FF0000} +0.065} &
  {\color[HTML]{FF0000} -0.055} &
  {\color[HTML]{FF0000} +4.83} &
  {\color[HTML]{FF0000} +0.058} &
  {\color[HTML]{FF0000} -0.056} \\ \hline
\end{tabular}%
}
\vspace{-4mm}
\label{tab:nvs-quant}%
\end{table}

\subsection{Photorealistic Rendering} 
We validate the photorealistic rendering performance of our simulator by evaluating image reconstruction and novel view synthesis (NVS) following ~\cite{ost_neural_2021,turki_suds_2023}. 

\ssec{Baselines} We conduct qualitative and quantitative comparisons against other state-of-the-art methods: NeRF~\cite{mildenhall_nerf_2020}, NeRF with timestamp input (denoted as NeRF+Time), NSG~\cite{ost_neural_2021}, PNF~\cite{kundu_panoptic_2022}, and SUDS~\cite{turki_suds_2023}. Note that none of them simultaneously meet all three standards mentioned in Table.~\ref{tab:recon-quant}.

\ssec{Implementation Details} Our model is trained for 200,000 iterations with 4096 rays per batch, using RAdam as optimizers. The learning rate of the background node is assigned $1*10^{-3}$ decaying to $1*10^{-5}$, while that of $5*10^{-3}$ decaying to $1*10^{-5}$ in object nodes.

\begin{figure}[ht]
    \vspace{-7mm}
    \centering
    \includegraphics[width=0.75\textwidth]{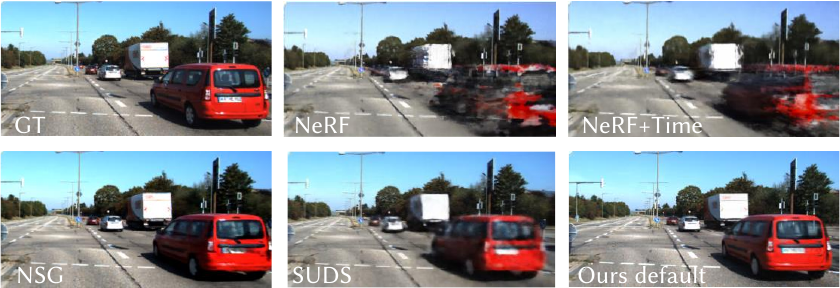}
    \caption{Qualitative image reconstruction results on KITTI dataset.}
    \label{fig:recon}
    \vspace{-7mm}
\end{figure}

\begin{figure}[ht]
    \centering
    \includegraphics[width=0.78\textwidth]{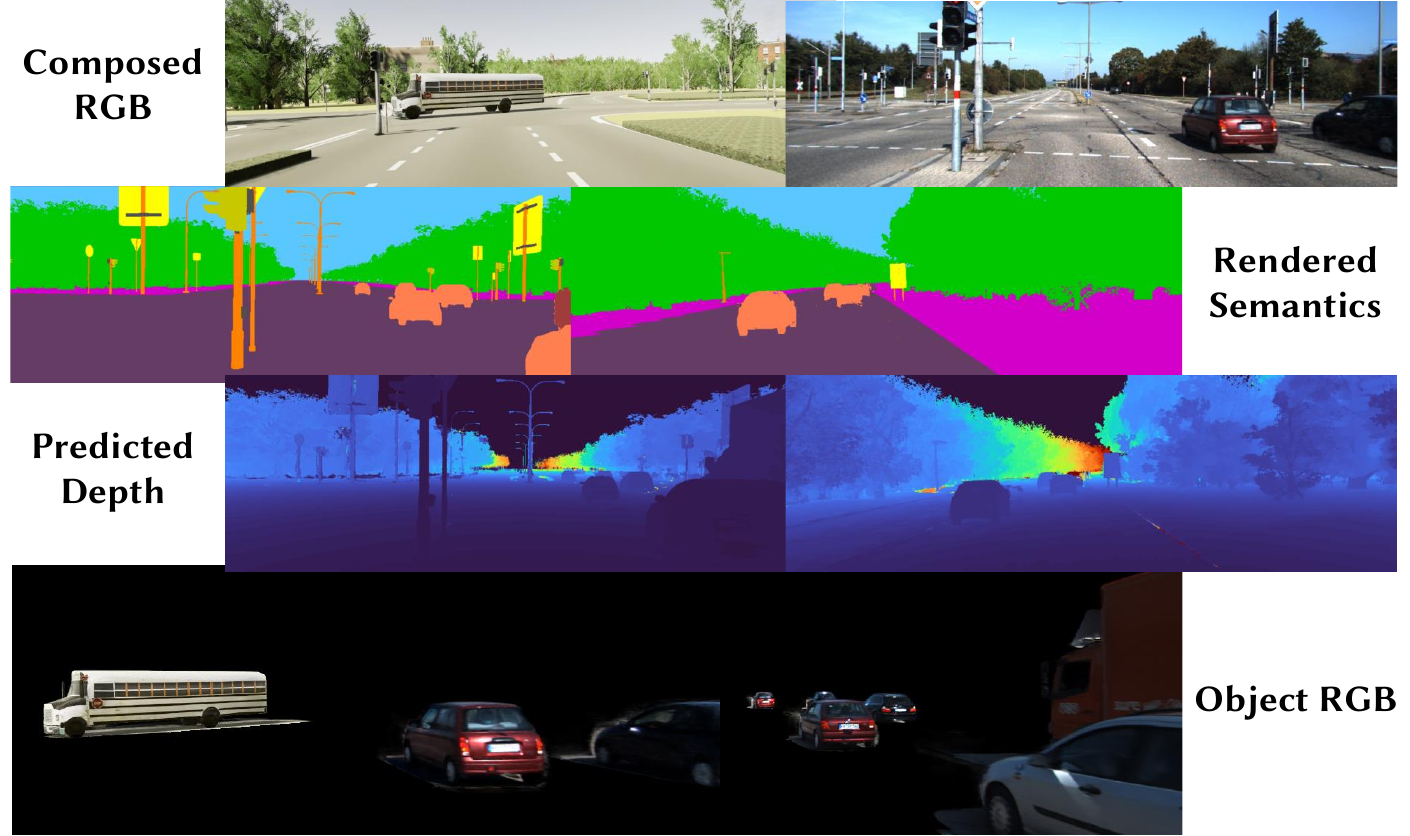}
    \caption{Gallery of different rendering channels.}
    \label{fig:qualitative}
    \vspace{-4mm}
\end{figure}

\ssec{Experiment Settings} The training and testing image sets in the image reconstruction setting are identical, while in the NVS task, we render the frames that are not included in the training data. Specifically, we hold out every 4$^\text{th}$ frames, every 2$^\text{nd}$ and 4$^\text{th}$ frames, and training with only one in every four frames, namely 25\%, 50\%, and 75\%. 

We follow the standard evaluation protocol in image synthesis and report Peak Signal-to-Noise Ratio (PSNR), Structural Similarity (SSIM), and Learned Perceptual Image Patch Similarity (LPIPS) ~\cite{zhang_unreasonable_2018} of our default setting for quantitative evaluations. Results are shown in Table~\ref{tab:recon-quant} for image reconstruction and Table~\ref{tab:nvs-quant} for NVS, which indicate that our method outperforms baseline methods in both settings. We can achieve 29.79 PSNR on V-KITTI using 75\% training data, while the best result previously published is 23.87.

\subsection{Instance-wise Editing} 
\begin{figure}[ht]
\vspace{-4mm}
    \centering
    \includegraphics[width=0.74\textwidth]{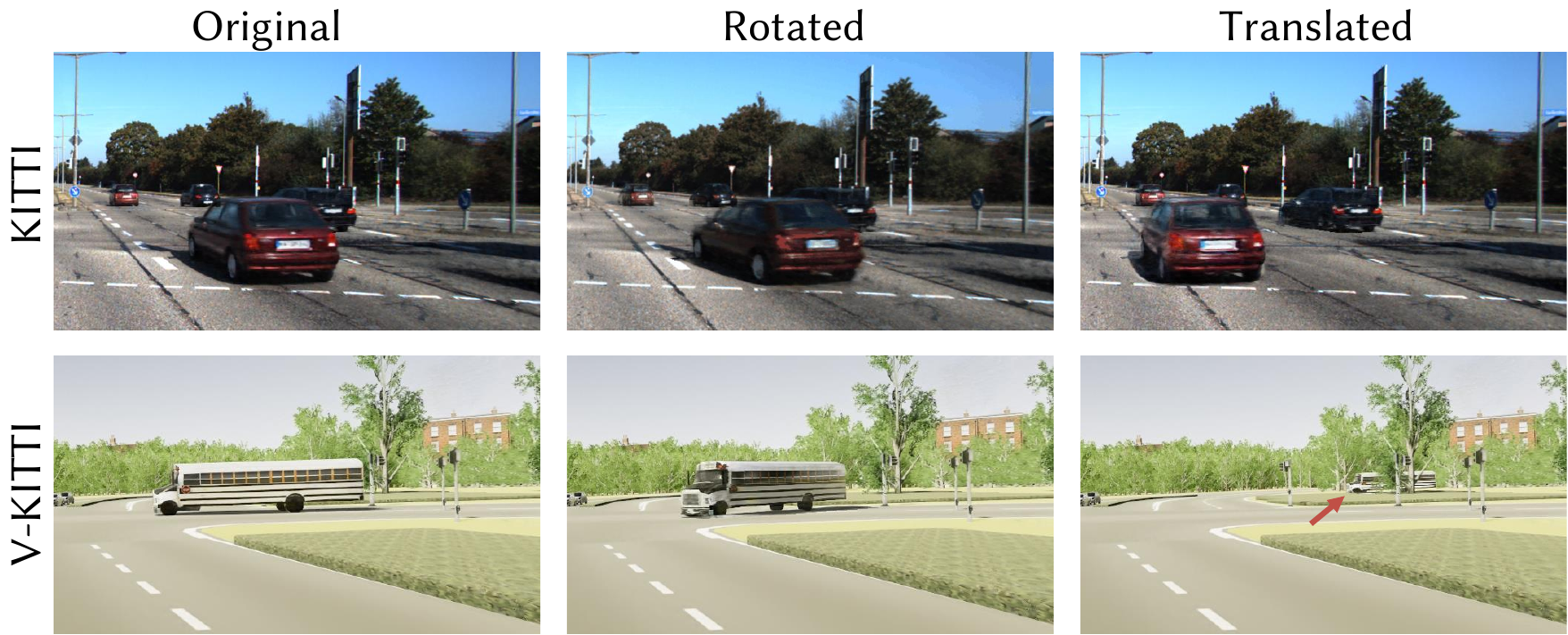}
    \vspace{-2mm}
    \caption{Rendering results on the edited scene.}
    \label{fig:editing}
    \vspace{-7mm}
\end{figure}

Our framework separately models background and foreground nodes, which allows us to edit the scene in an instance-aware manner. We qualitatively present our capability to remove instances, add new instances, and edit vehicle trajectories. In Fig.~\ref{fig:editing}, we show some editing examples of rotating and translating a vehicle, though more results can be found in our video clip.

\subsection{The blessing of moduler design}

We use different combinations of background and foreground nodes, samplers, and supervision signals for evaluation, which is credited to our modular design. 

Note that some of the baseline methods in the literature actually correspond to an ablation entry in this table. For instance, PNF ~\cite{kundu_panoptic_2022} uses NeRF as background node representation and instance-wise NeRF as foreground node representation with semantic losses. NSG~\cite{ost_neural_2021} uses NeRF as background node representation and category-level NeRF as foreground representation, but with a multi-plane sampling strategy. Our default setting uses grid-based background node representation, and our proposed category-level method for foreground node representation.




\subsection{Ablation Results}
\label{sec:ablation}
In this section, we analyze different experiment settings, verifying the necessity of our design. We reveal the impact of different design choices in background node representation, foreground node representation, etc. Specifically, we present all experiments with 50,000 iterations. Unlike prior works~\cite{turki_suds_2023,kundu_panoptic_2022,ost_neural_2021} that evaluate their method on a short sequence of 90 images, we use the full sequence from the dataset for all evaluation. Since they are not open-sourced and their exact evaluation sequences are not known, we hope our new benchmarking would standardize this important field. Quantitative evaluation can be found in Table~\ref{tab:ablation}.

For background and foreground nodes, we substitute our default model (ID 1 in Table~\ref{tab:ablation}) with MLP-based and grid-based model and list their metrics in row 2, 7-12. In the 3$^\text{rd}$-6$^\text{th}$ row, we show the effectiveness of our model components. For model and sampler, selected modules for background and foreground nodes are noted before and after the slash, respectively.

\begin{table}
\vspace{-5mm}
\centering
\caption{Quantitative evaluation for ablation studies}
\resizebox{0.85\textwidth}{!}{%
\begin{threeparttable}

\begin{tabular}{cccccccc|cccccc}
\hline
 &
  \multicolumn{7}{c|}{Settings} &
  \multicolumn{3}{c}{KITTI} &
  \multicolumn{3}{c}{V-KITTI} \\ \cline{2-14} 
\multirow{-2}{*}{ID} &
  Model &
  Sampler &
  Category &
  $\mathcal{L}_\text{sky}$ &
  \multicolumn{1}{c}{$\mathcal{L}_\text{depth}$} &
  $\mathcal{L}_\text{sem}$ &
  $\mathcal{L}_\texttt{accum}$ &
  PSNR $\uparrow$&
  SSIM $\uparrow$&
  LPIPS $\downarrow$&
  PSNR $\uparrow$&
  SSIM $\uparrow$&
  LPIPS $\downarrow$\\ \hline
\rowcolor[HTML]{EFEFEF} 
1\tnote{*} &
\multicolumn{1}{l}{\cellcolor[HTML]{EFEFEF}Grid / Ours} &
  \cellcolor[HTML]{EFEFEF}prop / c2f \tnote{$\dagger$} &
  \cellcolor[HTML]{EFEFEF} &
   &
   &
   &
   &
  \textbf{25.04} &
  \textbf{0.782} &
  \textbf{0.175} &
  \textbf{28.37} &
  \textbf{0.907} &
  \textbf{0.108} \\
2 &
  \textbf{MLP} / Ours &
  \textbf{c2f} / c2f &
   &
   &
  \multicolumn{1}{c}{} &
   &
   &
  20.14 &
  0.589 &
  0.476 &
  22.19 &
  0.664 &
  0.409 \\
\rowcolor[HTML]{EFEFEF} 
3 &
  Grid /   Ours &
  \cellcolor[HTML]{EFEFEF}prop / c2f &
  \cellcolor[HTML]{EFEFEF} &
   &
   &
   &
  × &
  21.35 &
  0.713 &
  0.242 &
  27.30 &
  0.881 &
  0.130 \\
4 &
  Grid /   Ours &
  prop / c2f &
   &
  × &
   &
  × &
   &
  23.68 &
  0.774 &
  0.181 &
  27.32 &
  0.881 &
  0.129 \\
\rowcolor[HTML]{EFEFEF} 
5 &
  Grid /   Ours &
  \cellcolor[HTML]{EFEFEF}prop / c2f &
  \cellcolor[HTML]{EFEFEF} &
   &
  \multicolumn{1}{c}{\cellcolor[HTML]{EFEFEF}×} &
   &
  \multicolumn{1}{l|}{\cellcolor[HTML]{EFEFEF}} &
  23.66 &
  0.769 &
  0.184 &
  27.30 &
  0.880 &
  0.128 \\
6 &
  Grid /   Ours &
  prop / c2f &
   &
  × &
  \multicolumn{1}{c}{} &
   &
  \multicolumn{1}{l|}{} &
  20.07 &
  0.723 &
  0.251 &
  27.42 &
  0.863 &
  0.148 \\
\rowcolor[HTML]{EFEFEF} 
7 &
  Grid / \textbf{MLP} &
  \cellcolor[HTML]{EFEFEF}prop / \textbf{c2f} &
  \cellcolor[HTML]{EFEFEF} &
   &
   &
   &
   &
  20.46 &
  0.709 &
  0.255 &
  26.46 &
  0.875 &
  0.132 \\
8 &
  Grid / \textbf{Grid} &
  prop / \textbf{prop} &
   &
   &
   &
   &
   &
  22.23 &
  0.741 &
  0.211 &
  25.22 &
  0.871 &
  0.134 \\
\rowcolor[HTML]{EFEFEF} 
9 &
  Grid / \textbf{MLP} &
  \cellcolor[HTML]{EFEFEF}prop / \textbf{c2f} &
  \cellcolor[HTML]{EFEFEF}× &
   &
   &
   &
   &
  20.98 &
  0.699 &
  0.257 &
  27.27 &
  0.881 &
  0.130 \\
10 &
  Grid / \textbf{Grid} &
  prop / \textbf{prop} &
  × &
   &
   &
   &
   &
  23.71 &
  0.763 &
  0.193 &
  26.65 &
  0.882 &
  0.125 \\ 
\rowcolor[HTML]{EFEFEF} 
11\tnote{*} &
  \textbf{MLP} / \textbf{MLP} &
  \cellcolor[HTML]{EFEFEF}\textbf{c2f} / \textbf{c2f} &
  &
  &
  &
  &
  &
  20.42 &
  0.592 &
  0.472 &
  21.77 &
  0.659 &
  0.410 \\
  \hline
\end{tabular}%
\begin{tablenotes}
    \footnotesize
    \item[$\dagger$] prop stands for proposal sampler, and c2f stands for coarse-to-fine sampler.
    \item[*] ID 1 is our default setting. ID 11 is similar to the setting of NSG \cite{ost_neural_2021} with coarse-to-fine sampler instead.
\end{tablenotes}    
\end{threeparttable}
}
\label{tab:ablation}
\vspace{-8mm}
\end{table}




\section{Conclusion}
In this paper, we present a modular framework for photorealistic autonomous driving simulation based on NeRFs. Our \textbf{open-sourced} framework consists of a background node and multiple foreground nodes, enabling the modeling of complex dynamic scenes.
We demonstrate the effectiveness of our framework through extensive experiments. The proposed pipeline achieved state-of-the-art rendering performance on public benchmarks. We also support different design choices of scene representations and sampling strategies, offering flexibility and versatility in the simulation process.

\vspace{1mm}\textbf{Limitations.} Our method requires hours to train and is not capable of rendering in real-time. Besides, our method fails to consider the dynamic specular effects on glasses or other reflective materials that may cause artifacts in rendered images. Improving simulation efficiency and view-dependent effects will be our future work.

\clearpage
%
%
%
\bibliographystyle{splncs04}
\bibliography{wuzirui_zotero}
\appendix
\section{Related Works}
\subsection{Neural Simulators}
Neural rendering has shown immense potential in representing autonomous driving scenarios. Some methods have been proposed to leverage this technology in simulators. 
Neural scene graph (NSG) \cite{ost_neural_2021} decomposes dynamic scenes by encoding object transformations into a learned scene graph, and learns latent representation to describe similar objects with the same implicit field. 
Panoptic Neural Fields (PNF) \cite{fu_panoptic_2022} went one step further to model instance-aware objects and jointly learn panoptic segmentation of the scene. 
Urban radiance field (URF) \cite{rematas_urban_2022} leverages predicted image segmentations to supervise densities on rays pointing at the sky. 
S-NeRF \cite{xie_s-nerf_2023} employs a camera transformation for object alignment and integrates reprojection and geometry confidence to enhance geometric consistency.
SUDS \cite{turki_suds_2023} factorizes the scene into three separate hash tables to efficiently encode static, dynamic, and far-field parts of the scene.
However, these methods failed to integrate the foreground and background well and leverage prior information about vehicles to create realistic scene representations in autonomous driving scenarios. Furthermore, the highly interconnected frameworks of these methods have limited their practical implementation in engineering projects. 

\subsection{Neural Scene Representations}
Neural Scene Representation has shown impressive results in image reconstruction and novel view synthesis. Neural Radiance Field(NeRF)\cite{mildenhall_nerf_2020} is the first to use implicit fields to represent scenes, various types of NeRFs have been proposed for acceleration. \cite{muller_instant_2022,tancik_nerfstudio_2023,chen_tensorf_2022} and better rendering quality \cite{barron_mip-nerf_2021,barron_mip-nerf_2022,wang_f2-nerf_2023}. NeRFStudio\cite{tancik_nerfstudio_2023} provides a unified framework for many of these approaches. However, most of them fail to reconstruct dynamic scenes with high quality, and there is no such framework available that offers a flexible module switch to seamlessly combine the distinct strengths of these approaches.

\section{More on the depth supervision}
We follow DS-NeRF \cite{deng_depth-supervised_2022} and utilize a ray distribution loss if the dense depth maps are available:
\begin{equation}
    \mathcal{L}_
    \text{sensor\_depth} = \sum_k\log h_k\exp\left(-\frac{(t_k-\mathcal{D}(\textbf{r}))}{2\hat{\sigma}_i^2}\right)\delta_k,
\end{equation}
where $\hat{\sigma}_i$ is the average reprojection error.

If the ground truth dense depth maps are not available, we use the monocular depth estimation network and employ a mono-depth loss \cite{yu_monosdf_2022} to regularize the geometry:
\begin{equation}
    \mathcal{L}_\text{mono\_depth} = \sum_\textbf{r}||w\hat{\textbf{d}}(\textbf{r}) + q - \mathcal{D}(\textbf{r})||_2^2,
\end{equation}
where $w$ and $q$ are the learnable scale and shift factors that are optimized through training for scale and shift-invariance.

\section{Details for foreground nodes}

\begin{figure}
\includegraphics[width=\textwidth]{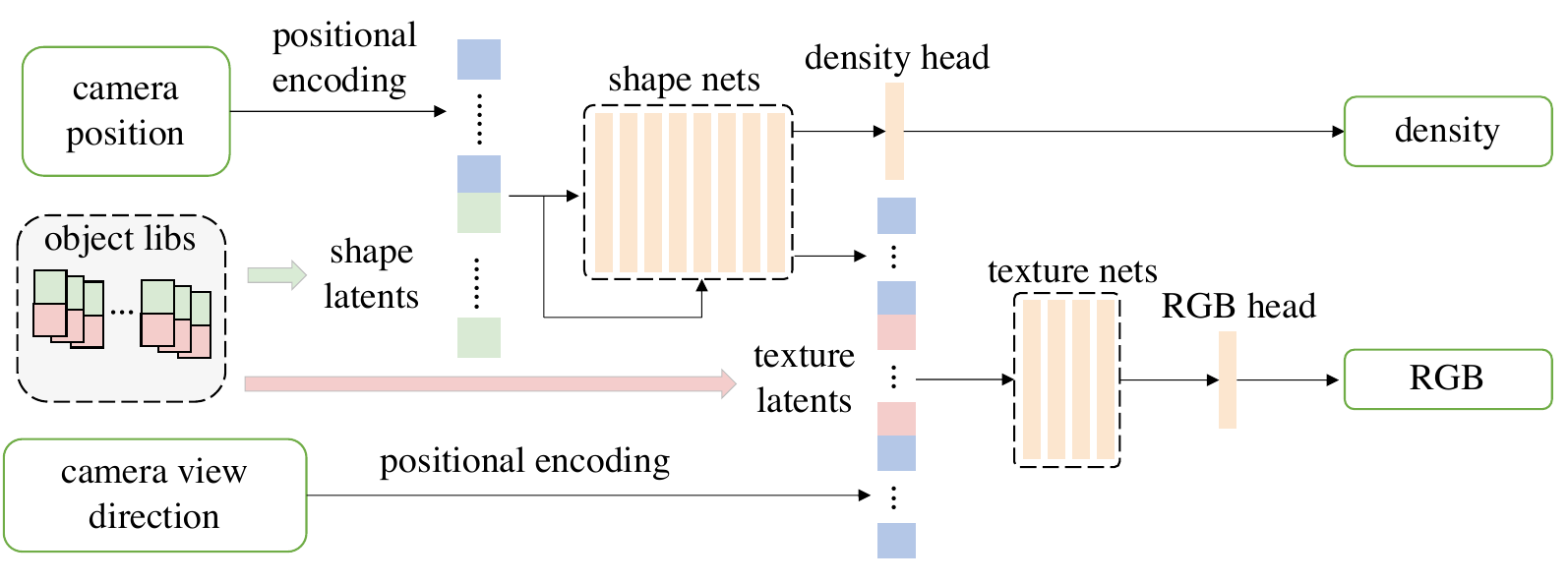}
\caption{Our decoupling shape and texture category-level model.} 
\label{fig:latents}
\end{figure}

The overview of our improved foreground nodes is shown in Fig.\ref{fig:latents}. Our approach learns the shape and texture of each object instance separately by decoupling them into two parts of latent codes. We optimize these latent codes over the test sequence and store them in object libraries. The object library is depicted in Figure \ref{fig:latents}.

During rendering, we incorporate the priors that are queried from the object library into the NeRF model. The shape net is an 8-layer MLP, and the texture net is a 4-layer MLP following the vanilla NeRF\cite{mildenhall_nerf_2020}. We combine the latents and the model by concatenating the positional encoded \cite{mildenhall_nerf_2020} camera position and the shape latent codes to model the instance's shape with the shape networks. The output feature is then stacked with the positional encoded viewing directions $\textbf{d}$ and the texture latents, and passed the resulting input to the texture networks to obtain the instance's texture.  These improvements contribute to \textbf{state-of-the-art} performance.

\begin{figure}
\includegraphics[width=\textwidth]{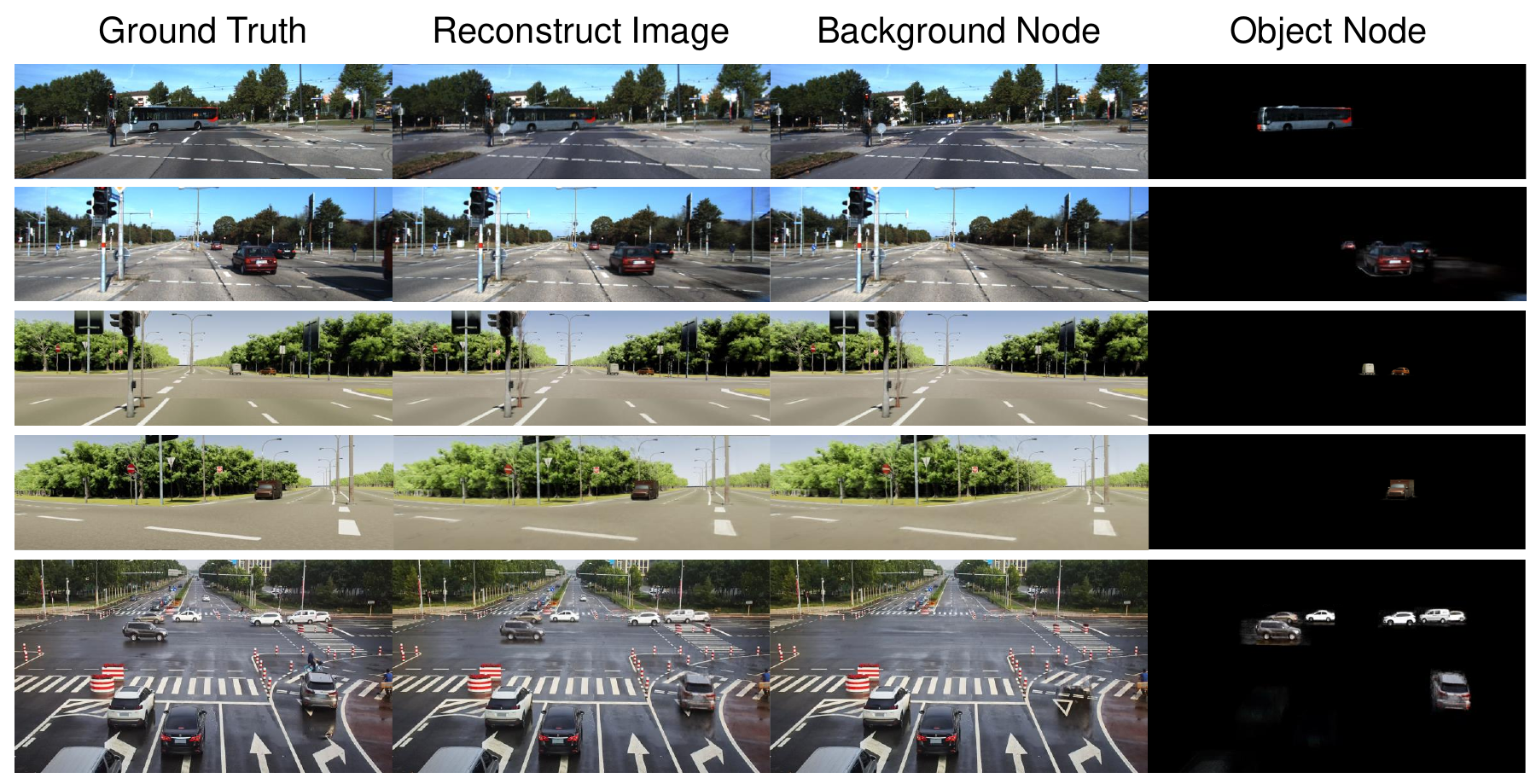}
\caption{Additional qualitative results demonstrating traffic scene decomposition effectiveness.} 
\label{fig:decomposition}
\end{figure}

\begin{figure}
\includegraphics[width=\textwidth]{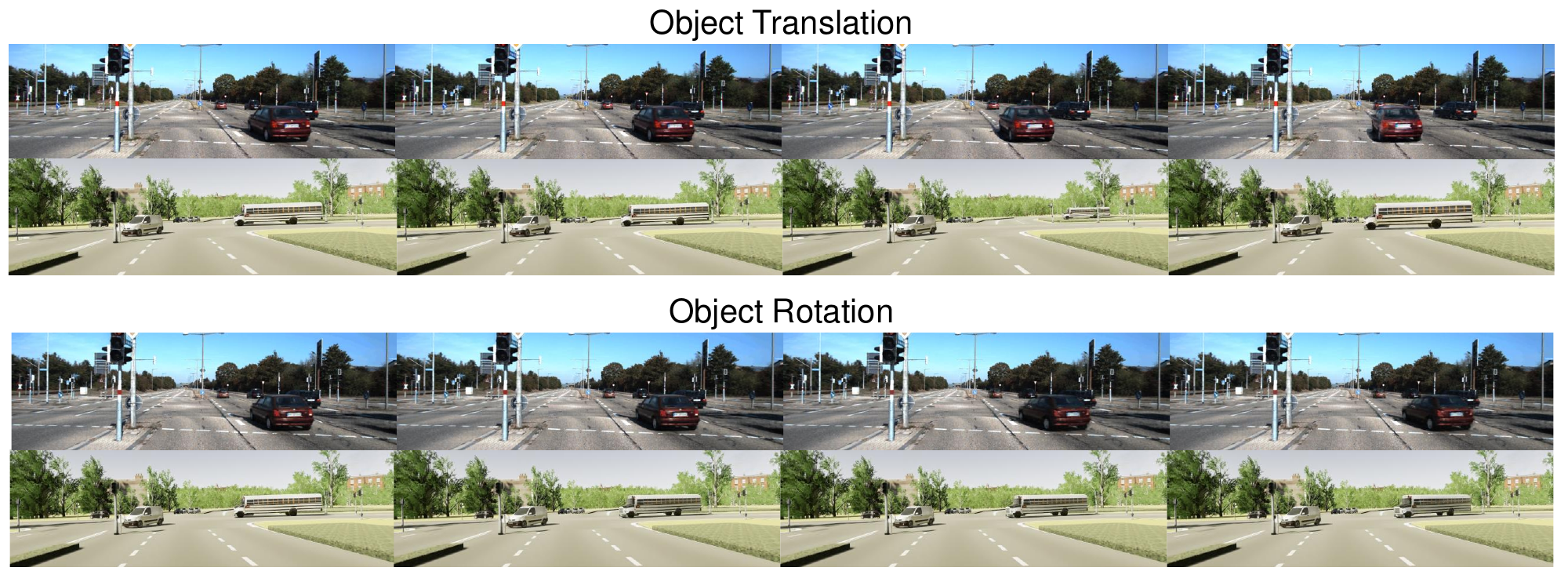}
\caption{Our method enables the generation of novel traffic environments by editing specific objects within a scene. In the first and third rows, the red sedan is edited, while in the second and fourth rows, the white bus is edited. } 
\label{fig:edit}
\end{figure}

\section{Synthetic traffic environment generation}

We present additional experiment results in Fig.~\ref{fig:decomposition} of decomposed rendering results on KITTI\cite{geiger_are_2012}, VKITTI\cite{cabon_virtual_2020}, and DAIR-V2X datasets \cite{yu_dair-v2x_2022}. Our method offers the benefits of background and foreground decomposition, which enables flexible editing of foreground nodes to generate novel synthetic traffic environments. Figure \ref{fig:edit} showcases the editing process, where the car can be deleted and added to a new position or orientation. These operations are equivalent to translating or rotating the target objects. These capabilities allow for the creation of diverse and customizable photorealistic traffic scenarios, which can be valuable for testing and evaluating autonomous driving systems and other traffic-related applications.

\end{document}